\documentclass{iosart2c}     

\usepackage[T1]{fontenc}
\usepackage{times}%
\usepackage{url}
\usepackage{graphicx}

\pubyear{2010}
\volume{1}
\firstpage{1}
\lastpage{9}

\begin{document}

\begin{frontmatter}    

\title{Optimizing real-time RDF data streams}
\runningtitle{Optimizing real-time RDF data streams}

\maketitle

\author{\fnms{Joshua} \snm{Shinavier}}
\address{Rensselaer Polytechnic Institute, Troy, NY USA}
\runningauthor{Joshua Shinavier}


\begin{abstract}
The Resource Description Framework (RDF) provides a common data model for the integration of ``real-time'' social and sensor data streams with the Web and with each other.  While there exist numerous protocols and data formats for exchanging dynamic RDF data, or \textit{RDF updates}, these options should be examined carefully in order to enable a Semantic Web equivalent of the high-throughput, low-latency streams of typical Web 2.0, multimedia, and gaming applications.  This paper contains a brief survey of RDF update formats and a high-level discussion of both TCP and UDP-based transport protocols for updates.  Its main contribution is the experimental evaluation of a UDP-based architecture which serves as a real-world example of a high-performance RDF streaming application in an Internet-scale distributed environment.
\end{abstract}


\end{frontmatter}


\section{Introduction}

Streaming data is an increasingly important component of the World Wide Web environment.  Social networking APIs such as Twitter and Facebook provide continuous, high-volume feeds of user content and activities, supporting an entire ecosystem of ``real time'' applications.  Mobile devices serve as personal gateways for a wide variety of near-real-time sensor data.  There are good reasons to integrate real-time data sources both with static Web data and with each other, and Semantic Web technologies provide a potential platform for that integration.  For example, mapping real-time social data into common Semantic Web vocabularies \cite{shinavier2010real}\cite{mendes2010twarql} enables ``smarter'' real-time queries which draw upon the wealth of general-purpose knowledge contained in the Linking Open Data cloud.\footnote{\url{http://esw.w3.org/SweoIG/TaskForces/CommunityProjects/LinkingOpenData}}  Bridging the gap between sensor data and the symbolic space of the Semantic Web \cite{dietze2009bridging} opens the door to a semantic Internet of Things, while the combination of social network data with sensor data \cite{breslin2009integrating} promises more personalized and contextually aware real-time services.

The Resource Description Framework (RDF) provides a common data model in which to express and combine schema-friendly information from diverse sources.  Furthermore, various notions of \textit{RDF updates} or changesets permit the communication of dynamic changes to that data, such as the posting of a photo or the change of a user's geolocation.  Emerging technologies such as SPARQL 1.1\footnote{\url{http://www.w3.org/TR/2009/WD-sparql11-http-rdf-update-20091022/}} and sparqlPuSH \cite{passant-sparqlpush} provide transport mechanisms for updates.  As the Semantic Web moves into this new domain, then, performance and scalability issues should be kept in mind from the start.

The content of this paper is as follows. Section~\ref{rdf_update_formats} will survey currently available RDF update formats.  Section~\ref{transport_protocols} will discuss transport protocols for RDF update streams at a high level. Both TCP-based and the hitherto unexplored option of UDP-based update streams will be discussed.  Section~\ref{compression} will argue in favor of lossless data compression regardless of the choice of protocol.  Finally, Section~\ref{evaluation} will describe a concrete implementation of a distributed, UDP-based solution in which a volume of data equivalent to the Twitter Firehose is pushed from a client machine to a remote server\footnote{The scripts and programming notes for this research are open source and can be found here: \url{http://github.com/joshsh/laboratory/tree/master/research/rdfstream/}.}
\footnote{All numerical results were derived using the following hardware and software:
\begin{itemize}
\item \textit{sending machine}: Ubuntu 9.10 Server on an Amazon EC2 ``small'' virtual machine in Bloomsbury, NJ (USA) with 2GB RAM, 160 GB disk, and one single-core, 2.66 GHz Intel Xeon processor E5430
\item \textit{receiving machine}: Ubuntu Server 10.04 on a rack-mounted server in Oakland, CA (USA) with 64 GB RAM, 2TB disk, and eight 4-core, 2.13 GHz Intel Xeon processors E5506
\end{itemize}} and ingested into an RDF triple store in real time, such that the data is immediately available for query through a SPARQL endpoint.


\section{RDF update formats}
\label{rdf_update_formats}

For some applications, it may be sufficient to think of updates simply as streams of RDF triples.  A news feed, for example, may describe each new story as a distinct resource, neither replacing nor invalidating descriptions which have gone before.  In this case, an RDF update feed might be nothing more than a succession of RDF/XML documents, or perhaps SPARQL results.  Other applications, however, are more stateful.  A user's ``mood'' which changes from ``sad'' to ``happy'' is ambiguous if the addition of the new mood is not preceded by the deletion of the old one.  All of the RDF update formats described below support the addition and deletion of statements, while many of them also support further operations such as the creation of named graphs or the definition of namespaces.

For the moment, only the vocabulary component of these technologies will be considered.  We will also ignore the subtle distinction between \textit{change} formats, which express a difference between RDF graphs or successive states of an RDF graph, and \textit{update} formats, which apply an update operation to a graph database.

\subsection{SPARQL/Update}
    
The SPARQL/Update\footnote{\url{http://www.w3.org/TR/sparql11-update/}} \cite{seaborne2008sparql} language, nicknamed SPARUL, comes closest to a standard update language for RDF graphs.  Using a syntax derived from SPARQL, SPARUL provides several basic update operations, including statement-level insertion and deletion.
See Figure~\ref{sparul_example} for an example.

Also worth mentioning are the very similar SparqlUpdateLanguage\footnote{\url{http://esw.w3.org/SparqlUpdateLanguage}}, as well the SPARQL update syntax of ARC's SPARQL+\footnote{\url{http://arc.semsol.org/docs/v2/sparql+}}.

\begin{figure*}
\caption{SPARQL/Update example}
\label{sparul_example}
\begin{verbatim}
PREFIX dc: <http://purl.org/dc/terms/>
DELETE { <http://example.org/ns#resource1> dc:title "Original Title"  }
INSERT { <http://example.org/ns#resource1> dc:title "New Title" }
\end{verbatim}
\end{figure*}

\subsection{Delta ontology}

The Delta ontology\cite{berners2004delta} and Notation3\footnote{\url{http://www.w3.org/DesignIssues/Notation3}}-based file format apply the notions of textual \textit{diff} and \textit{patch} to RDF graphs, permitting the syndication of changes to graphs distributed among two or more peers.  See Figure~\ref{delta_example} for an example.

\begin{figure*}
\caption{Delta ontology example}
\label{delta_example}
\begin{verbatim}
@prefix diff: <http://www.w3.org/2004/delta#>.
@prefix dc: <http://purl.org/dc/terms/>.
{ <http://example.org/ns#resource1> dc:title "Original Title" }
    diff:replacement
{ <http://example.org/ns#resource1> dc:title "New Title" }.
\end{verbatim}
\end{figure*}

\subsection{Changesets}

Changesets\footnote{\url{http://n2.talis.com/wiki/Changesets}} is a resource-oriented scheme for tracking changes to an RDF graph.  An update, or changeset, is centered on a single \textit{subject of change}, such that the change is specific to the bnode closure, or \textit{concise bounded description}, of that resource.

The Changeset RDF vocabulary\footnote{\url{http://vocab.org/changeset/schema.html}} uses RDF reification to express changes in terms of triples added or removed, and additionally includes terms to express meta-information about a change, including its time and purpose, the entity responsible, and the preceding change in a history of changes. See Figure~\ref{changesets_example} for an example.

\begin{figure*}
\caption{Changesets example}
\label{changesets_example}
\begin{verbatim}
<rdf:RDF xmlns:rdf="http://www.w3.org/1999/02/22-rdf-syntax-ns#" 
    xmlns:cs="http://purl.org/vocab/changeset/schema#">
  <cs:ChangeSet rdf:about="http://example.org/changes#change1">
    <cs:subjectOfChange rdf:resource="http://example.org/ns#resource1"/>
    <cs:removal>
      <rdf:Statement>
        <rdf:subject rdf:resource="http://example.org/things#resource1"/>
        <rdf:predicate rdf:resource="http://purl.org/dc/terms/title"/>
        <rdf:object>Original Title</rdf:object>
      </rdf:Statement>
    </cs:removal>
    <cs:addition>
      <rdf:Statement>
        <rdf:subject rdf:resource="http://example.org/things#resource1"/>
        <rdf:predicate rdf:resource="http://purl.org/dc/terms/title"/>
        <rdf:object>New Title</rdf:object>
      </rdf:Statement>
    </cs:addition>
  </cs:ChangeSet>
</rdf:RDF>
\end{verbatim}
\end{figure*}

\subsection{GUO}

The Graph Update Ontology (GUO)\footnote{\url{http://webr3.org/specs/guo/}} defines an RDF diff in terms of triple-level insert and delete operations.  Like the Changesets vocabulary, GUO expresses an update as an RDF resource, allowing additional metadata to be attached to the update.  Unlike Changesets, GUO avoids RDF reification and supports named graphs.  See Figure~\ref{guo_example} for an example.

\begin{figure*}
\caption{GUO example}
\label{guo_example}
\begin{verbatim}
@prefix rdf: <http://www.w3.org/1999/02/22-rdf-syntax-ns#> .
@prefix dc: <http://purl.org/dc/terms/> .
@prefix guo: <http://webr3.org/owl/guo#> .

_:u0 rdf:type guo:UpdateInstruction ;
     guo:target_subject <http://example.org/ns#resource1> ;
     guo:delete _:d0 ;
     guo:insert _:i0 .
_:d0 dc:title "Original Title" .
_:i0 dc:title "New Title" .
\end{verbatim}
\end{figure*}

\subsection{GRUF}

The Guaranteed RDF Update Format (GRUF)\footnote{\url{http://websub.org/wiki/GRUF}} is a proposed plain-text format for RDF updates.  While there are currently no software implementations of GRUF, it is more compact than any of the other formats described here, making it potentially appropriate for high-volume RDF update streams.  It supports both triples and named graph quads.  See Figure~\ref{gruf_example} for an example.

\begin{figure*}
\caption{GRUF example}
\label{gruf_example}
\begin{verbatim}
set_subject http://example.org/things#resource1
set_property http://purl.org/dc/terms/title
delete text Original Title
add text New Title
\end{verbatim}
\end{figure*}

\subsection{Sesame RDF transactions}

The Sesame 2.0 RDF framework includes a document format for RDF updates which has been given the media type application/x-rdftransaction.  Statement-level \textit{add} and \textit{remove} operations are expressed with subject-predicate-object triple patterns which include an optional named graph component.  See Figure~\ref{rdftransaction_example} for an example.  The ease of parsing the simple XML-based format efficiently has led to its reuse in the AllegroGraph triple store,\footnote{\url{http://www.franz.com/agraph/allegrograph/doc/new-http-server.html#header2-235}} and consequently in the experimental evaluation described in Section~\ref{evaluation}.

\begin{figure*}
\caption{Sesame RDF transactions example}
\label{rdftransaction_example}
\begin{verbatim}
<transaction>
    <remove>
        <uri>http://example.org/things#resource1</uri>
        <uri>http://purl.org/dc/terms/title</uri>
        <literal>Original Title</literal>
        <contexts/>
    </remove>
    <add>
        <uri>http://example.org/things#resource1</uri>
        <uri>http://purl.org/dc/terms/title</uri>
        <literal>New Title</literal>
        <contexts/>
    </add>
</transaction>
\end{verbatim}
\end{figure*}

\subsection{Other formats}

Several other RDF update formats have been proposed, although they are closely tied with uncommon query languages.  The rdfDB Query Language\footnote{\url{http://www.guha.com/rdfdb/query.html#insert}}, for example, provides insert and delete operations. The Modification Exchange Language (MEL) \cite{nejdl2002towards} is based on RDF reification and interoperates with Edutella's\footnote{\url{http://www.edutella.org/edutella.shtml}} Query Exchange Language (QEL). The RDF Update Language (RUL) \cite{magiridou2005rul} deals with type-safe class and property instance level updates, interoperating with the query and view languages RQL and RVL.

Furthermore, there are many additional technologies which deal with RDF updates and change notification, without however providing a statement-level update format.  For example, the Triplify update vocabulary\footnote{\url{http://triplify.org/vocabulary/update}} alerts data consumers to incremental changes by providing pointers to RDF documents which have changed, while DSNotify\footnote{\url{http://dsnotify.org/}} and the Web of Data Link Maintenance Protocol\footnote{\url{http://www4.wiwiss.fu-berlin.de/bizer/silk/wodlmp/}} facilitate synchronization of Linked Data link sets.


\section{Transport protocols}
\label{transport_protocols}

Most of the RDF update formats described in the preceding section are intended to be used in conjunction with a particular communication protocol.  SPARQL/Update, for example, is now associated with the SPARQL 1.1 protocol for managing RDF graphs. The RDF transactions format is not even a proposed standard, having only been intended for use with Sesame's HTTP protocol.\footnote{\url{http://www.openrdf.org/doc/sesame2/system/ch08.html}} Changesets has its own HTTP-based protocol,\footnote{\url{http://n2.talis.com/wiki/Changeset_Protocol}} although it is also used simply as a vocabulary for representing changes.
SparqlPuSH, which embeds SPARQL query results in RSS and Atom feeds, uses the PubSubHubbub protocol\footnote{\url{http://code.google.com/p/pubsubhubbub/}} to proactively broadcast updates to data subscribers via HTTP POST.

Given the origins of the Semantic Web, is not surprising that nearly all of these protocols are based on HTTP.  The proposed XMPP bindings for the SPARQL protocol\footnote{\url{http://danbri.org/words/2008/02/11/278}} are an exception, while at a higher level, both HTTP and XMPP are usually layered upon the Transmission Control Protocol (TCP).  It is these lower levels of protocol which impose the most basic constraints on both the latency and throughput of RDF update streams sent over the Internet, so we will discuss them in the following, illustrating their well-known properties with small-scale experiments.

In somewhat more depth, we will also examine another core member of the Internet Protocol suite, the User Datagram Protocol (UDP), and explore the constraints it imposes as a carrier of RDF updates.

\subsection{Basic observations}

TCP is a reliable, connection-based protocol, which entails both advantages and disadvantages with respect to latency and throughput.  Establishment of a connection involves the overhead of an initial handshake, after which a two-way stream of bytes flows efficiently between endpoints.  Packets are guaranteed to arrive intact and in order.  However, this requires that any lost packets are retransmitted, incurring additional delays.  UDP, on the other hand, is connectionless and guarantees only that individual datagrams will either arrive intact or not at all, and that in indeterminate order.  It therefore avoids the overhead of an initial handshake and of retransmission of lost packets, at the expense of reliability.

Studies of TCP throughput for bulk transfer of data suggest that it is governed by 
a handful of factors including round-trip time and a path-specific probability of packet loss \cite{padhye1998modeling}.  As packet loss increases, throughput drops according to successively higher-order exponentials, making TCP increasingly inefficient over congested or otherwise lossy networks. UDP throughput, in contrast, drops off in direct proportion to packet loss.

\subsection{HTTP GET}

\begin{figure}
\caption{Data throughput using HTTP GET}
\label{http_get_throughput}
\includegraphics[scale=0.45]{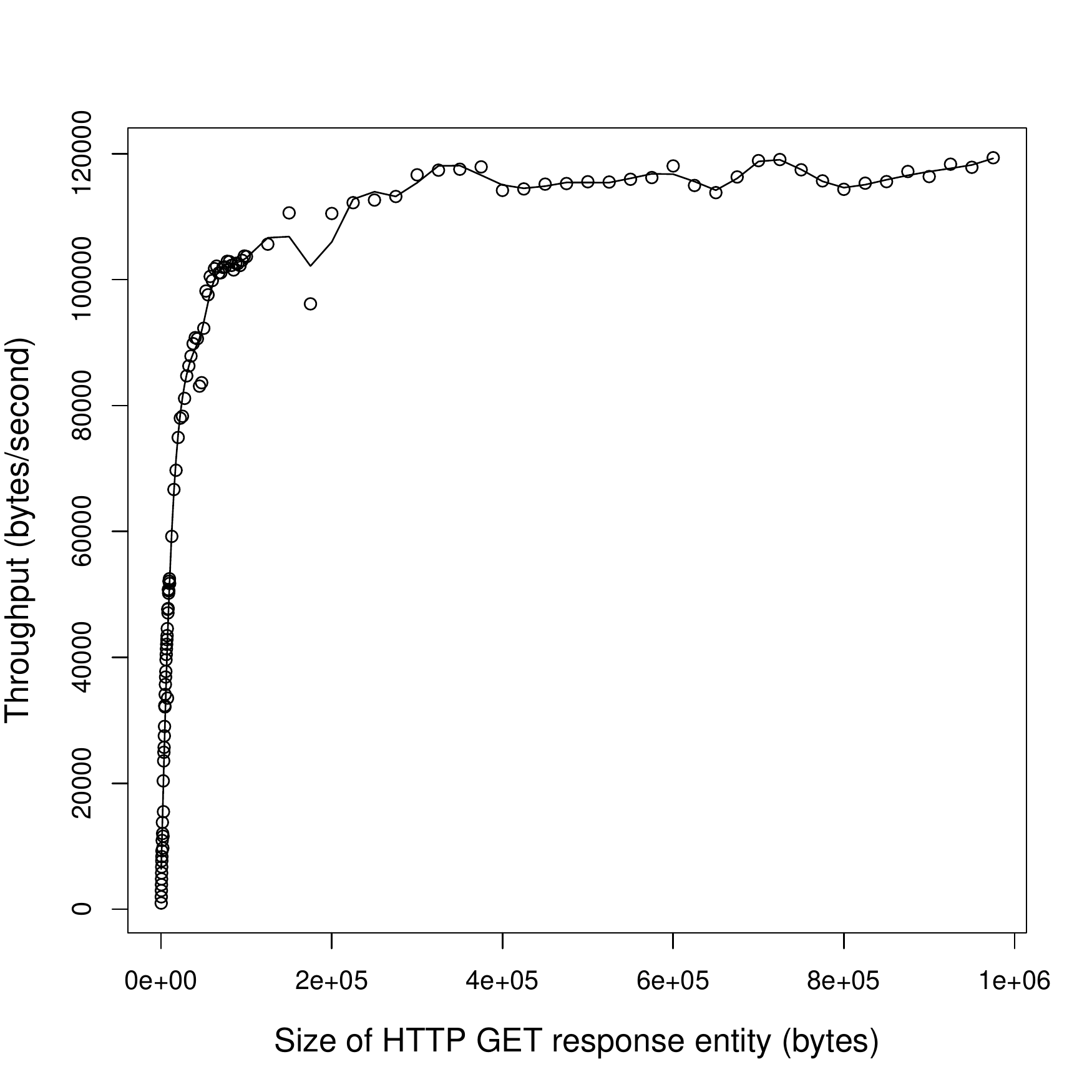}
\end{figure}

HTTP's GET method is primarily used to retrieve Web resources such as HTML pages, images, JavaScript documents, and so on, based on their URIs.  To do so, a client sends an HTTP request message to the server, which is met with an HTTP response which is transmitted to the client over a persistent connection.  Since the client does not need to re-negotiate the TCP connection after the initial request, a larger response body results in a higher proportion of data received to time spent.  This is illustrated in Figure~\ref{http_get_throughput}, in which a client has repeatedly retrieved a document of varying size from a server.\footnote{Each data point is based on 100 HTTP GET requests from a Java program on the sending machine to an AllegroServe HTTP server on the receiving machine.} If data is retrieved in 1,000-byte chunks, around 9 requests per second are possible in the experimental environment, or 9,000 bytes per second.  If, however, data is retrieved in 100,000-byte chunks, only one request per second is possible, but this amounts to over 10 times as much data per second.  In terms of throughput of RDF updates, successive large documents -- or a single, continuous stream -- are preferable to a larger number of smaller documents.  Therefore, it should be possible to group multiple updates into a single document or stream.

\subsection{HTTP POST}

\begin{figure}
\caption{Data throughput using HTTP POST}
\label{http_post_throughput}
\includegraphics[scale=0.45]{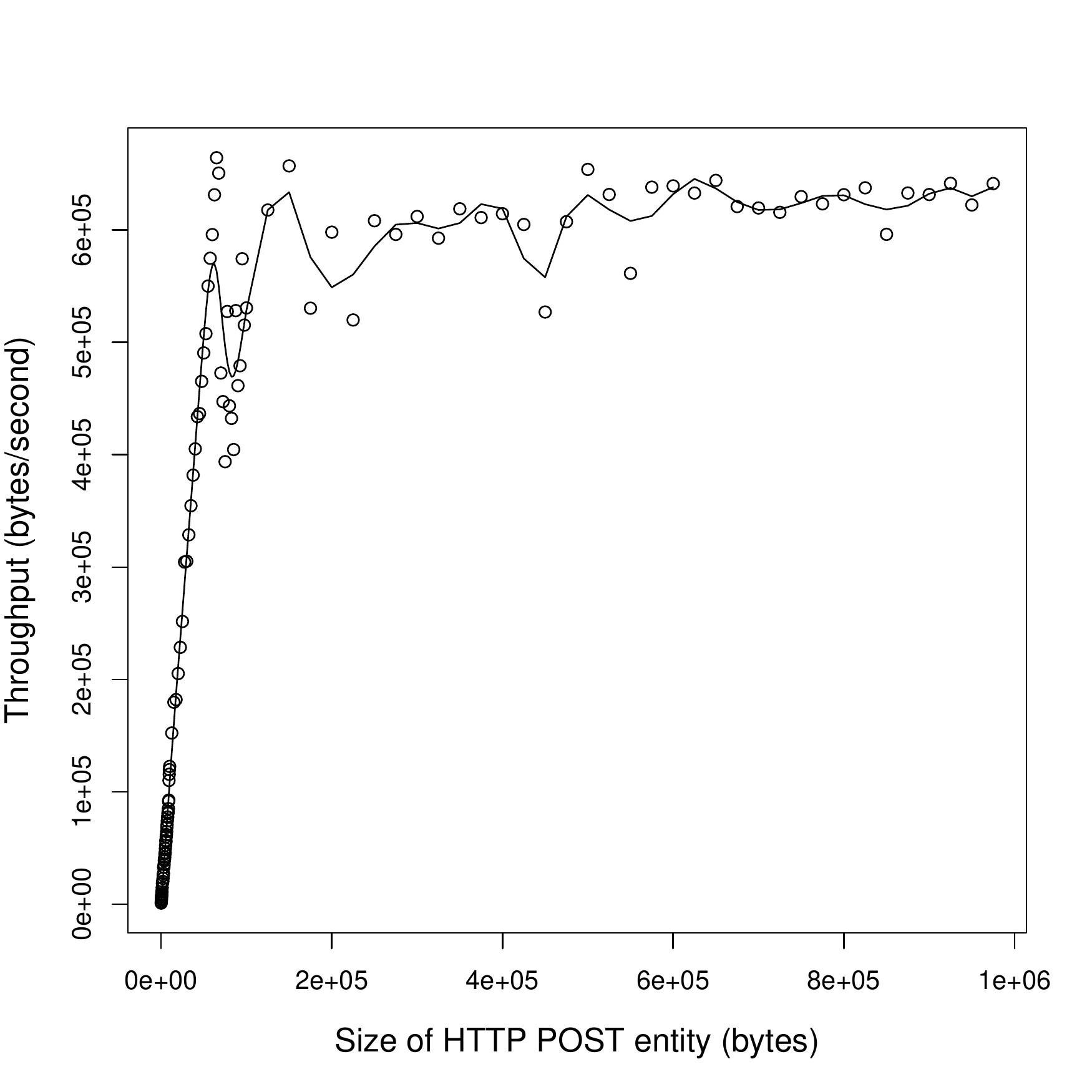}
\end{figure}

POST requests follow exactly the same pattern as GET responses (see Figure~\ref{http_post_throughput}):\footnote{Each data point is based on 100 POST requests from a Java program on the sending machine to an AllegroServe HTTP server on the receiving machine. The apparent difference in absolute throughput for GET and POST is due to the opposite direction of flow of the HTTP payload between the two machines, which are subject to a difference in download and upload bandwidth.} the larger the body of successive requests, the higher the throughput.  In analogy to GET, this is an argument in favor of update protocols which allow for multiple update operations per POST.

\subsection{UDP}

\begin{figure}
\caption{Data throughput using UDP}
\label{udp_throughput}
\includegraphics[scale=0.45]{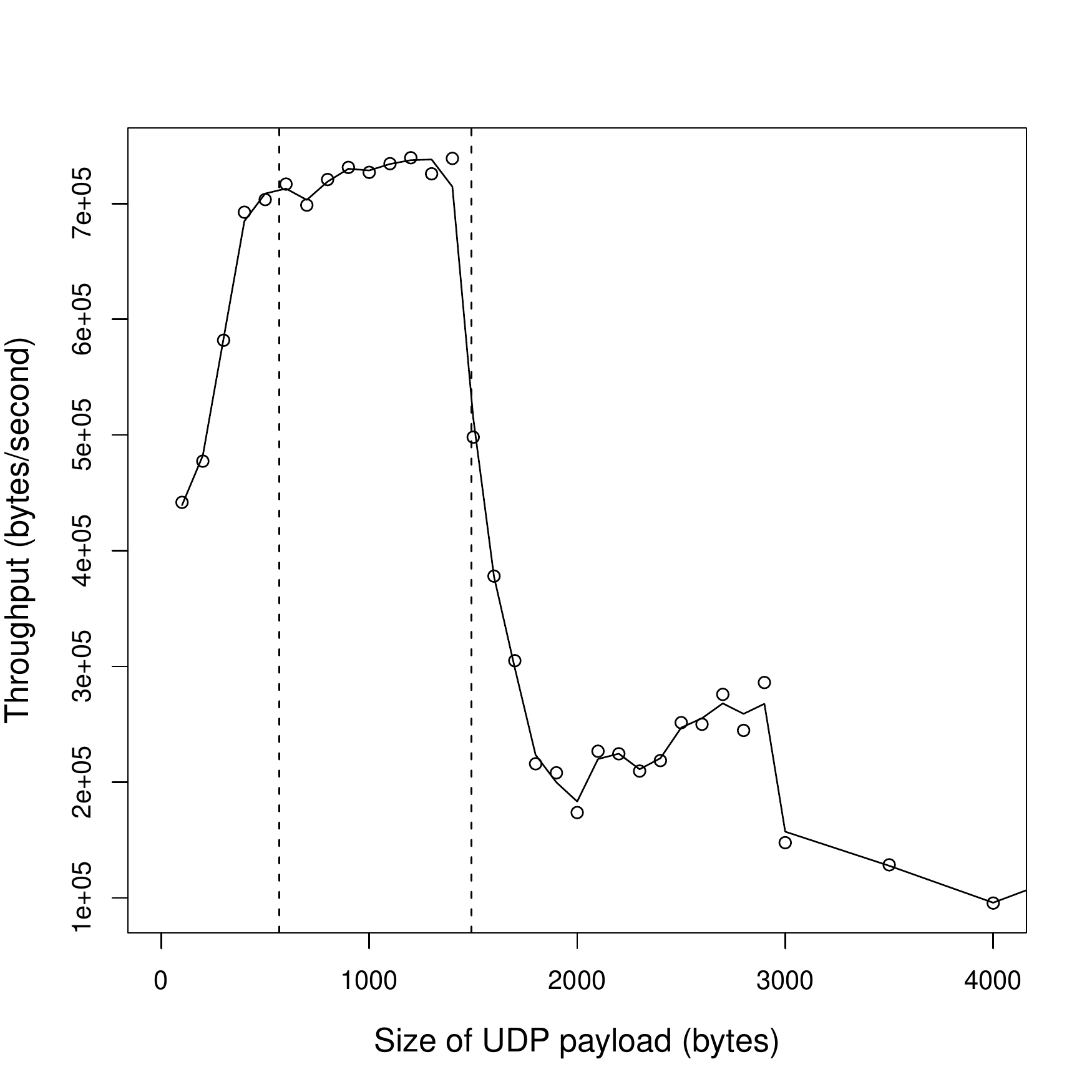}
\end{figure}

UDP throughput follows an altogether different pattern than that of TCP-based HTTP GET or POST (see Figure~\ref{udp_throughput}).\footnote{Each data point is based on 1000 UDP datagrams from a Java program on the sending machine to an Allegro Common Lisp program on the receiving machine.}  Since a UDP payload is contained in a single packet or \textit{datagram}, its size is limited by the maximum transmission unit (MTU) of the path.  Most of the Internet is subject to the Ethernet v2 frame format, which imposes an MTU of 1500 bytes.  As a UDP datagram includes a 8-byte header, the payload should be no larger than 1492 bytes (indicated in the figure with a dashed line).  All IP v4 hosts must be prepared to accept datagrams up to 576 bytes, so a UDP payload of less than 568 bytes (also indicated with a dashed line) is sub-optimal.  Datagrams larger than the MTU are subject to fragmentation, with negative implications for throughput.  The second rising slope in the figure is evidence of datagrams which have been broken into two fragments each.

This constraint, together with the unreliability of UDP, imposes several requirements on a streaming RDF application:
\begin{enumerate}
\item data loss, to some extent, is acceptable
\item it is possible to break updates into small, atomic transactions, such that the loss of  individual transactions will not corrupt the RDF database on the receiving end
\item communication is one-way, such the sender does not require acknowledgement of receipt of transactions
\item order of delivery of transactions is not important
\end{enumerate}

UDP-based update streams are therefore not a general-purpose solution, but they may confer advantages, in terms of latency and throughput, for certain very demanding applications.  For example, the framework described in Section~\ref{evaluation} addresses a use case in which there is more data than available bandwidth and the main concern is to transmit as high a proportion of the data as possible.  For another example, some varieties of sensor data are so time-sensitive that it is better to drop lost updates than to attempt to retransmit them, particularly when sensors operate under less-than-perfect network conditions. Similarly, the frequent use of UDP for online multiplayer games hints at use cases for UDP-based RDF streams in real-time interactive environments, with potential applications in pervasive computing and augmented reality.

In addition, UDP allows the possibility of \textit{IP multicasting}, in which RDF updates are broadcast from a single data producer to a practically unlimited number of data consumers.



\section{Compression}
\label{compression}

Lossless compression of updates is found to be beneficial in all cases.  However, the choice of a compression format, as well as efficient implementations of the compression and decompression algorithm, are relevant.  Figure~\ref{compression} illustrates the effect of three lossless compression strategies on the size of a small RDF update document in the Sesame RDF transaction format.\footnote{Based on a sample set of 100,000 tweets}  Beginning with an average message of over 5,000 bytes, the document is reduced to the target size of a UDP datagram by two of the strategies, both of which support fast compression and decompression.  

Implementations of DEFLATE in Java and Common Lisp were found to both compress and decompress RDF transaction documents of 100,000 bytes or less in under a millisecond each, making the compression overhead significantly less than the corresponding gain in throughput.

\begin{figure}
\caption{Comparison of compression formats}
\label{compression}
\includegraphics[scale=0.45]{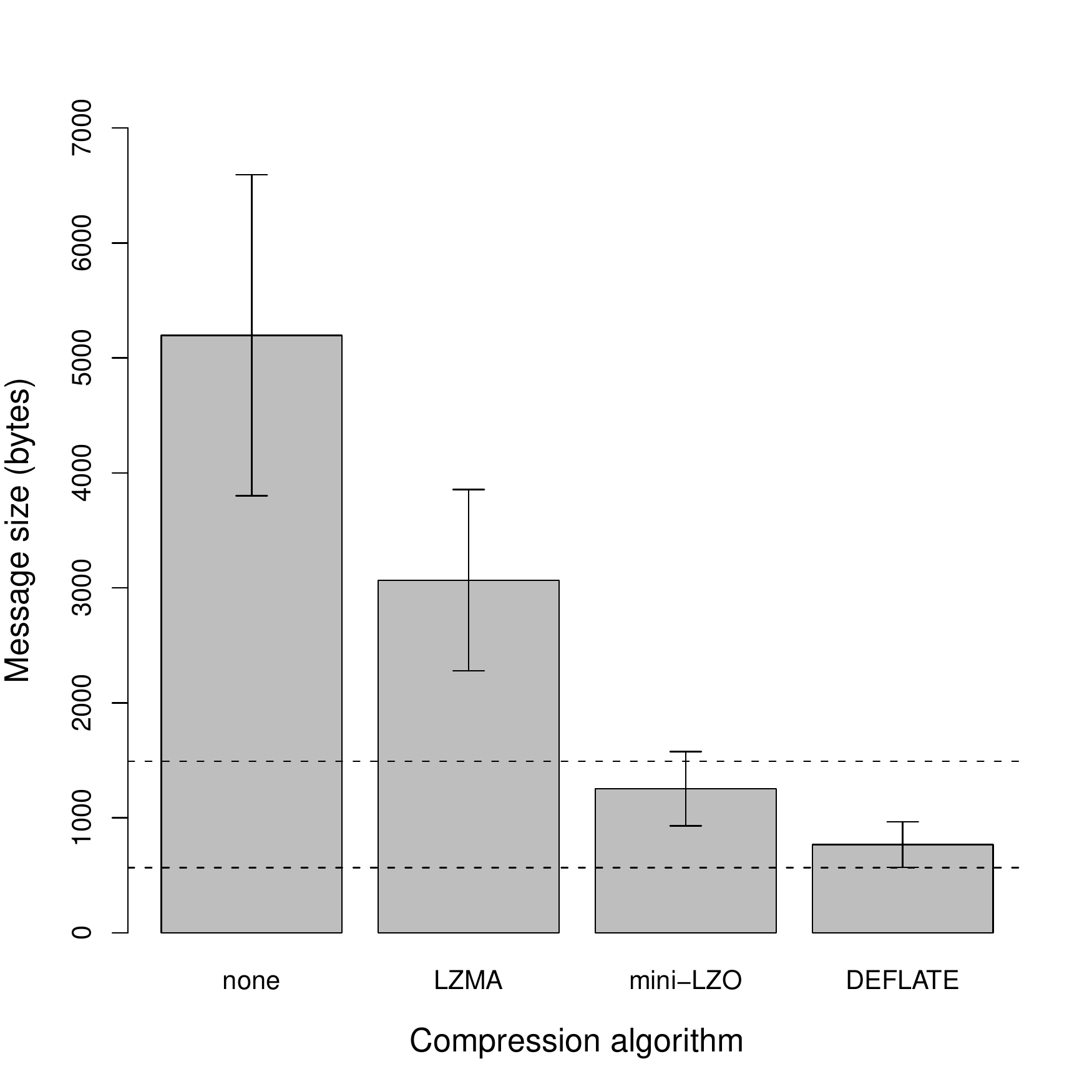}
\end{figure}


\section{Implementation and evaluation}
\label{evaluation}

The framework described in this section\footnote{Source code for this study is available at: \url{http://fortytwo.net/research/rdfstream}} is motivated by large-scale social networking services which generate more data than it is possible to transport between a single pair of widely separated Internet hosts.  Surprisingly, by giving up on transporting \textit{all} of the data, we are in fact able to transport \textit{more} data than would otherwise be possible.

\subsection{RDFizing the Twitter Firehose}

Twitter's Streaming API\footnote{\url{http://dev.twitter.com/pages/streaming_api}} provides near-real-time access to various subsets of public and protected Twitter data.  The most privileged level of access, the Firehose stream, contains some 500 to 1000 status updates, or ``tweets'' per second,\footnote{according to community estimates} where a tweet consists of a short snippet of text accompanied by several dozen fields of metadata including time and place of the update and a description of the author.

When translated into RDF using TwitLogic,\footnote{\url{http://twitlogic.fortytwo.net}} a tweet is  represented with an average of 18 triples.\footnote{Based on a sample set of 100,000 tweets}  In order to update the triple store appropriately (for example, replacing the tweet author's location with a new location), an average of 9 ``remove'' operations are also required, for a total of just under 30 update operations per tweet.  In this experiment, we simulate the Twitter Firehose by generating a high volume of randomized tweets, translating them from Twitter JSON to individual RDF transactions in the Sesame RDF transaction format.

\subsection{Transporting RDF transactions}

In order to validate the UDP-based technique described above, all RDF transactions were transported from the sending machine to the receiving machine in individual UDP datagrams, having first been DEFLATE-compressed.  Datagrams were dispatched at a rate of over 1000 per second, without regard to whether they were received.

The receiving machine processed successfully transmitted datagrams as quickly as possible, expanding and parsing the RDF transaction payload of each message, which is immediately applied to an AllegroGraph triple store, pending commit.

\subsection{Persistence}

A transactional database requires a ``commit'' operation to permanently apply previously executed update operations.  In the case of AllegroGraph, a commit is relatively expensive, costing at least 4 milliseconds regardless of the number of update operations to commit.  Committing multiple tweets at a time was found to be more efficient, with a ratio of 100 RDF transactions, or 3000 update operations to 1 commit practically minimizing commit overhead.

At that point, a single Lisp client on the receiving machine was found to process between 128 and 269 transactions per second,\footnote{all results were computed with an initially empty AllegroGraph triple store which grew to a size of 65 million triples by the end of the experiment. The results were not found to be affected significantly by the addition of 240 million triples of DBpedia data} depending on the status of the background merge process: immediately after merging indices, the triple store accepts over twice as many transactions per second as it does immediately before completion of the merge.  Write performance then degrades at a roughly constant rate as new transactions are committed, repeating in a sawtooth pattern for successive merge cycles.  Without the use of a transaction buffer, it is the minimum write performance which defines the triple store's ability to keep up with an incoming stream of data.

\subsection{Multiprocessing}

\begin{figure}
\caption{Multithreaded data ingest}
\label{ag_multithreading}
\includegraphics[scale=0.45]{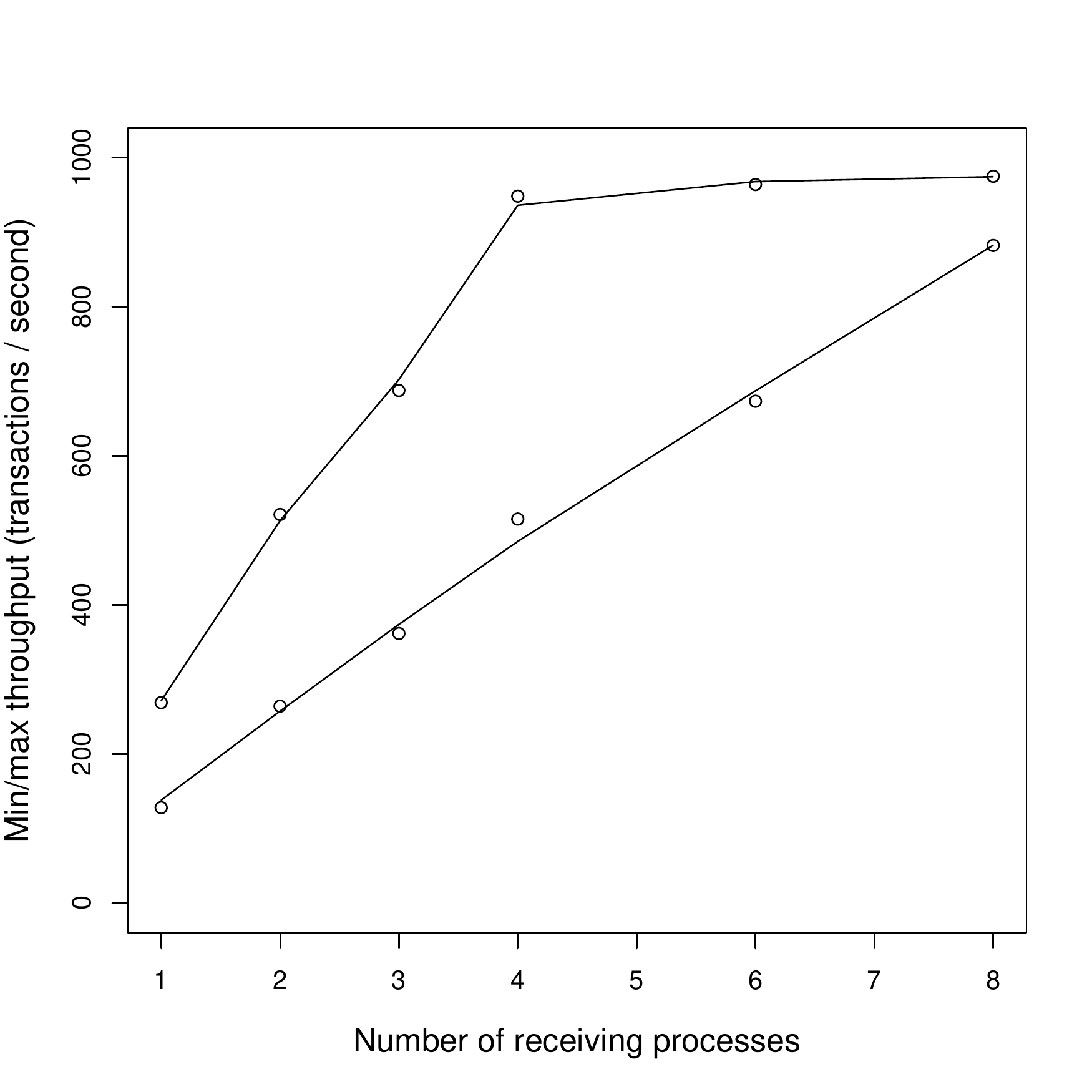}
\end{figure}

In AllegroGraph, data ingest can be facilitated by making use of multiple triple store clients, each in its own Lisp-based process.  As shown in Figure~\ref{ag_multithreading}, data throughput then increases nearly linearly as new clients are created, sharing the load of data ingest.  This experiment was performed with 1, 2, 3, 4, 6 and 8 client processes, where each client receives UDP messages on a separate port and the sending machine distributes outgoing messages evenly across those ports.  The upper line in the figure represents the combined maximum throughput -- immediately after a merge -- of all receiving processes.  At four processes, maximum throughput begins to level off at around 980 transactions per second, or 98\% of the throughput of the sending machine.  In other words, messages are consumed as quickly as they are produced, disregarding packet loss.  At eight threads, minimum throughput is around 90\% of the ceiling value, or 88\% of the total stream.

Overall, the system successfully commits around 930 tweets per second to the remote triple store, which is close to the estimated volume of the Twitter Firehose.

\subsection{Possibilities for scalable query answering}

In the above, we have demonstrated a low-latency, high-throughput solution for streaming RDF updates and data ingest.  In order to make good use of this data, however, real-time query capabilities are also required.  This presents a scalability challenge if data ingest places high computational demands on the ingesting machine.\footnote{This is not necessarily the case in the above, as we did not make full use of the multiprocessing capability of the receiving machine.}  In AllegroGraph, a file-based transaction log offers the possibility of replicating a primary triple store on any number of secondary machines, which then share the burden of query answering among themselves.  This functionality has been implemented but has yet to be tested in a high-throughput setting such as the above.  It requires the overhead of reading from the transaction log to be less than the overhead of receiving RDF transactions over UDP.  Otherwise, IP multicasting provides a better solution.


\section{Conclusion}

In the above, we have surveyed two of the most important technical choices surrounding general-purpose, real-time RDF data streams -- namely, data formats and transport protocols -- with an eye towards maximizing data throughput.  In the case of formats, there is no shortage of options, which may be distinguished from one another in terms of their relative compactness, ease of generation and parsing, and the presence of mature implementations.  Although most of these formats are associated with individual HTTP-based protocols, throughput is limited primarily by a handful of factors common to all of them, including message size and the use of data compression.  In particular, protocols which make use of HTTP POST can dramatically increase their performance ceiling by sending an arbitrary number of atoms of data per connection.  Given current tools, lossless compression always confers a performance advantage.

In addition to HTTP POST and GET, we have also considered an alternative, UDP-based technique for RDF data streaming.  We have argued that it offers a slight performance advantage as a replacement for high-volume HTTP-based streams, but that it may be most appropriate in future real-time Semantic Web applications for which minimal latency is the overriding concern.

Finally, we have illustrated our observations with a real system which implements the UDP-based technique, evaluating its performance with respect to an oft-cited example of a high-volume data stream, the Twitter Firehose.  This system, which combines an RDF triplification tool with an RDF update stream and an RDF graph database, is presented as evidence that current Semantic Web technologies are up to the task of participating in highly demanding real-time applications.


\section{Future work}

Although the update formats and transport protocols surveyed above serve as a starting point for the development of high-performance RDF streaming applications, there are many more possibilities to be explored.  For example, the Datagram Congestion Control Protocol (DCCP) and the Stream Control Transmission Protocol (SCTP) are both message-oriented protocols which add congestion control, MTU discovery, and a measure of reliability over UDP.  Alternatively, IP multicasting may prove useful in the broadcasting of RDF updates by popular real-time data providers.  Finally, it is worth noting that the goal achieved in the preceding section -- that of transporting, ingesting and querying over the Twitter Firehose -- is a reasonably well-defined yet rather informal benchmark with respect to throughput of RDF data.  Much as concrete metrics such as the Lehigh University Benchmark (LUBM)\footnote{\url{http://swat.cse.lehigh.edu/projects/lubm/}} have been developed to evaluate RDF graph databases in terms of integrity and scalability, so there is also a need for metrics which address throughput and response time of highly dynamic real-time Semantic Web services.


\section{Acknowledgements}

This work has been supported by Franz Inc.\footnote{\url{http://www.franz.com/}} as well as Rensselaer Polytechnic Institute's Tetherless World Constellation.  Special thanks go to Jans Aasman and Marko A. Rodriguez, among many others, for their helpful feedback.


\bibliographystyle{amsplain}
\bibliography{rdfstream}

\end{document}